\documentclass{article}

\usepackage{PRIMEarxiv}

\usepackage[utf8]{inputenc} 
\usepackage[T1]{fontenc}    
\usepackage{hyperref}       
\usepackage{url}            
\usepackage{booktabs}       
\usepackage{amsfonts}       
\usepackage{nicefrac}       
\usepackage{microtype}      
\usepackage{lipsum}
\usepackage{fancyhdr}       
\usepackage{graphicx}       

\usepackage{balance} 
\usepackage{multirow}
\usepackage{listings}
\usepackage{subfigure}
\usepackage{subcaption}
\usepackage{mathtools}
\usepackage{makecell}

\usepackage{soul}  
\usepackage{xcolor}
\usepackage[colorinlistoftodos]{todonotes}
\usepackage[export]{adjustbox}
\usepackage{cleveref}

\usepackage{algorithm}
\usepackage[noend]{algorithmic}
\usepackage[numbers]{natbib}

\newtheorem{definition}{Definition}

\newcommand{\fbits}{$\texttt{FBI}_\texttt{LTL}$}
\newcommand{\fbitsnaive}{$\texttt{FBI}_\texttt{LTL}^\texttt{naive}$}

\graphicspath{{media/}}     

\title{Diverse Planning with Simulators via Linear Temporal Logic
}

\author{
  Mustafa F. Abdelwahed, Alice Toniolo, Joan Espasa, Ian P. Gent \\
  School of Computer Science \\
  University of St Andrews \\
  United Kingdom\\
  \texttt{\{ma342, a.toniolo, jea20, Ian.Gent\}@st-andrews.ac.uk} \\
}

\begin{document}
\maketitle

\begin{abstract}

Autonomous agents rely on automated planning algorithms to achieve their objectives. Simulation-based planning offers a significant advantage over declarative models in modelling complex environments. However, relying solely on a planner that produces a single plan may not be practical, as the generated plans may not always satisfy the agent's preferences. To address this limitation, we introduce \fbits{}, a diverse planner explicitly designed for simulation-based planning problems. \fbits{} utilises Linear Temporal Logic (LTL) to define semantic diversity criteria, enabling agents to specify what constitutes meaningfully different plans. By integrating these LTL-based diversity models directly into the search process, \fbits{} ensures the generation of semantically diverse plans, addressing a critical limitation of existing diverse planning approaches that may produce syntactically different but semantically identical solutions. Extensive evaluations on various benchmarks consistently demonstrate that \fbits{} generates more diverse plans compared to a baseline approach. 
This work establishes the feasibility of semantically-guided diverse planning in simulation-based environments, paving the way for innovative approaches in realistic, non-symbolic domains where traditional model-based approaches fail.

\end{abstract}


\section{Introduction}

Autonomous agents rely on automated planning algorithms to achieve their objectives~\cite{automated-planning-agents}. These agents usually employ a planner that generates a single plan and executes it. However, the generated plan may not consider an agent's preferences because it was not accounted for during modelling the problem. 
A suitable solution is to employ a diverse planner that produces a set of diverse plans that account for various factors. Subsequently, the agent can select the plan that better aligns with its preferences. The benefits of using diverse planners can range from anticipating potential future situations~\cite{haessler1991cutting} to accommodating side information, such as preferences, which can be challenging to model~\cite{nguyen2012generating}.


%
Diverse planners are widely used in various real-world applications. For instance, they are employed in risk management~\cite{sohrabi2018ai}, malware detection~\cite{boddy2005course,sohrabi2013hypothesis}, and business process automation~\cite{chakraborti2020robotic}.
Typically, these planners use a symbolic representation of the planning problem to find plans. This representation defines the actions a planner can apply to a specific state based on the action’s preconditions. This representation is constructed using declarative languages like STRIPS~\cite{fikes1971strips} and PDDL~\cite{aeronautiques1998pddl}. However, this approach faces a challenge when modelling actions with complex effects that are difficult to represent using such languages. Planning with simulators emerges as an alternative~\cite{lipovetzky2015classical}. In this approach, the declarative model is replaced with a simulator that accepts a state and an action and generates a new state, called a successor state, as a result of applying the action to the input state.
Applying diverse planning in these simulation-based problems would offer the advantage of providing different solutions for an agent to pick from. For instance, diverse planners for simulators can be beneficial for revealing several vulnerabilities in a network through physically executing cyber-attacks on the network under test. Another scenario that would benefit from diverse planners for simulators is for playability of video games. \citet{game-diversity-measure} proposed using agent (e.g., A$^\star$ agent) to play the game to produce several traces (i.e., a sequence of actions that solves the level). Then use the generated traces to compute the diversity of the game. This is a crucial aspect that significantly influences player engagement~\cite{game-diversity-measure}, since the more diverse the game is the more engaging it is to the player. 


Generally diverse planners are model-based, requiring a symbolic representation of the planning problem. To our knowledge, in the literature the only approach to generate diverse plans for planning with simulators is presented by \citet{benke2023diverse}. They generated diverse plans by extracting them from pre-generated trees. Such a technique occasionally extracts similar plans when considered semantically~\cite{benke2023diverse} (i.e., doing the same task but with a different set of actions). \citeauthor{benke2023diverse} modelled diversity using a distance function which is not expressive enough to account for an agent's individual perspective on what makes two plans different. \citet{abdelwahed2024behaviour} proposed a framework called behaviour planning that aims to generate diverse plans based on a provided diversity model. A diversity model is represented using a set of features. Different from existing systems, \citet{abdelwahed2024behaviour} demonstrates how plans can be differentiated on the basis of this customisable model of diversity, which can account for an agent's individual preferences and goals. This framework can overcome the limitation of \citeauthor{benke2023diverse}’s approach in generating diverse plans for simulation-based planning problems.
%
%
\citeauthor{abdelwahed2024behaviour} implemented behaviour planning using the planning-as-satisfiability approach, but this implementation requires a declarative model for the planning problem, which is not the case here. 

In this paper we present an implementation of the behaviour planning framework \cite{abdelwahed2024behaviour} to generate diverse plans for agent planning with simulators. 
Since benchmark datasets for planning with simulator problems are currently unavailable~\cite{benke2023diverse}, we study the advantages of our approach using two real-world scenarios  for diverse planning with simulators. The first problem is a retro game called Puzznic, and the second is a network penetration planning problem.
We further evaluated our planner on a set of benchmark problems in PDDLGym~\cite{pddlgym}. The results demonstrate that our planner can produce diverse plans for planning with simulator problems. 

This paper presents two contributions:  a novel implementation of a behaviour planning framework for  simulators and a set of simulation-based planning problems used as benchmark to evaluate our system. 
The paper is structured as follows: first, we provide the necessary background knowledge. Then, we cover the implementation of our planner. Next, we present the experiments and their discussions. Finally, we delve into the related work on diverse planning with simulators and conclude the paper by outlining future work.

\section{Background}\label{sec:background}

At a high level, our behaviour planning implementation generates diverse plans for agents using simulation-based planners based on a given diversity model. This is achieved by considering a diversity model during the plan generation phase. This section delves into the essential topics for understanding our behaviour planning implementation. First, we introduce the diversity planning problem, and the diversity representation used; then, we highlight the elements of the behaviour planning framework. Subsequently, we cover the Iterated Width (\texttt{IW(i)}) planner used for behaviour planning and Linear Temporal Logic (\texttt{LTL}) for the  diversity model representation used by behaviour planning.

\textbf{Diversity planning problems.} Following \citet{ghallab2016automated}, a planning problem is a tuple $\Xi=\langle S, A, \gamma, \operatorname{cost}, I, G\rangle$, where $S$ is a set of states, $A$ is a set of actions, and $\gamma: S \times A \rightarrow S$ is a transition function that associates each state $s\in S$ and action $a\in A$ to the next state $\gamma(s,a)=s^\prime$. The function $\operatorname{cost}: A \rightarrow \mathbb{R}^+$ represents the cost of an action, and we only consider costs associated to actions $cost(a)$ irrespective of the state in which they were applied. $I\in S$ represents the initial state, and $G\in S$ is the goal formula. 
A solution for $\Xi$ is a plan ($\pi$) defined as a sequence of actions $a_1, a_2, \ldots, a_m$ such that $a_i\in A$ and $\gamma(\gamma(\gamma(I, a_0),\ldots),a_m) =G$. The cost of a plan $\pi$ is computed by accumulating the costs of its actions, such that $\operatorname{cost}(\pi)=\sum_{a_i\in\pi}cost(a_i)$. We overload the notation of $\operatorname{cost}$ for simplification. The set of all plans that solve $\Xi$ is referred to as $\Pi_\Xi$. In this work, we factor any state into a set of Boolean predicates.

In this work, we employ the diversity planning problem formulation proposed by \citet{abdelwahed2024behaviour} which represented diversity using an n-dimensional grid that is inspired from the Diversity-Quality optimisation field~\cite{lehman2011abandoning}, where each dimension represents a feature of interest. This grid is called \emph{behaviour space}. Any box within this grid is referred to as a \emph{behaviour}.  Diversity is quantified by counting the number of behaviours covered by a given set of plans, in a metric called the \emph{behaviour count}. 
Formally, the behaviour space ($BS_{\Delta_\Xi}= \Delta_1 \times \Delta_2 \times \ldots \times \Delta_n$) formed by dimensions $\Delta_\Xi$ is constructed using a set of user-defined features $F_\Xi=\{f_1,\ldots,f_n\}$. A feature $f_i=\langle \Delta_i, \odot_i\rangle$ includes a domain $\Delta_i$ containing the values of feature $f_i$ and $\odot:\Pi_\Xi\rightarrow \Delta_i$ an extracting function that extracts the value of feature $f_i$ from a given plan. All extracting functions are grouped into a set called $\odot_\Delta$.  A plan behaviour $\mathcal{B}=\langle\delta^1,\ldots,\delta^n\rangle$ is an $n$-tuple containing a value $\delta^i\in\Delta_i$ for each dimension $\Delta_i\in\Delta_\Xi$. To calculate the behaviour count, given a plan $\pi$ and a set of extracting functions $\odot_{\Delta}$, the plan behaviour is extracted by $\operatorname{PBehaviour}(\odot_\Delta, \pi)=\langle \odot_1(\pi), \odot_2(\pi),\ldots,\odot_n(\pi)\rangle$. The behaviour count is then $BC(\odot_\Delta,\Psi_\Xi)=\left| \{ \operatorname{PBehaviour}(\odot_\Delta, \pi) \mid \pi \in \Psi_\Xi \} \right|$.  
The formulation of the diversity problem is defined as follows:


\begin{definition}[diversity planning problem]\label{def:planning-task}
    Given a planning problem $\Xi$, a cost bound $c$, a number of plans $k$ and a set of dimensions $\Delta_\Xi$, find a set of plans $\Psi_\Xi\subseteq\Pi_\Xi$ that is subject to $cost(\pi) \leq c\ \forall \pi\in\Psi_\Xi$, $|\Psi_\Xi|\leq k$ and the behaviour count $BC(\odot_\Delta,\Psi_\Xi)$ of $\Psi_\Xi$ should be maximised. 
\end{definition}

 \citet{abdelwahed2024behaviour} used \citet{sreedharan2020emerging} personas to describe the users of behaviour planning. The personas involved in the planning system are the end user, who interacts with the system through a user interface; the domain designer, who sets high-level mission objectives; and the algorithm designer, who generates plans based on the end user and domain designer's requirements. In this work, we consider the agent is the end user that would select plans that align with their preferences and the domain domain expert is the one responsible for implementing the behaviour space's dimensions.
 




\textbf{Behaviour planning.} Following \citet{abdelwahed2024behaviour}, this is formed by two components: the first component is a diversity modelling approach called \emph{Behaviour Sorts Suite} (\texttt{BSS}). The second component is a planning approach called \emph{Forbid Behaviour $\text{\emph{Iterative}}_\texttt{X}$} ($\texttt{FBI}_\texttt{X}$), which generates diverse plans based on the \texttt{BSS}’s represented diversity model.
\texttt{BSS} represents diversity models (i.e., behaviour spaces) as defined above. 
After a behaviour space $BS_{\Delta_\Xi}$, $BS_\Delta$ for short, is constructed for a planning task $\Xi$, Forbid Behaviour $\text{Iterative}_\texttt{X}$ ($\texttt{FBI}_\texttt{X}$) uses this behaviour space to generate diverse plans. $\texttt{FBI}_\texttt{X}$ finds a plan, forbids its behaviour and then keeps doing this loop until all valid behaviours are generated or the number of required plans is achieved. \Cref{alg:fbi-planner-main} outlines the primary operation of the $\texttt{FBI}_\texttt{X}$ planner. It commences with an empty set of plans ($\Psi_\Xi$) (Line \ref{alg-line:bspace-planner-initial-set}), and subsequently, it iteratively generates and accumulates plans incorporating novel behaviours until either the $\operatorname{BehaviourGenerator}_\texttt{X}$ function ceases to produce plans or the specified limit ($k$) is reached (Lines \ref{alg-line:fbi-main-loop-start}-\ref{alg-line:fbi-main-loop-end}). 
\citeauthor{abdelwahed2024behaviour} suggested an extension to \Cref{alg:fbi-planner-main} to deal with cases where the number of required plans is greater than the number of available behaviours (i.e., $k > \vert BS_\Xi\vert$). This extension starts with invoking \Cref{alg:fbi-planner-main} to generate a plan for each behaviour and collects those plans, then resumes the same loop as $\texttt{FBI}_\texttt{X}$ except that it generates plans instead of behaviours using another function called $\operatorname{PlanGenerator}_\texttt{X}$. 


\Cref{alg:fbi-planner-main} is a general algorithm that generates diverse plans based on a set of provided diversity features. In this work, we propose a  modified version of the Iterative Width  Planner that supports planning with simulators to implement $\operatorname{BehaviourGenerator}_\texttt{X}$, and we chose  \texttt{LTL} to represent the diverse behaviour of plans for its wide use and expressiveness.

\begin{algorithm}
    \caption{$\operatorname{FBI}_\texttt{X}$ \cite{abdelwahed2024behaviour}}  \label{alg:fbi-planner-main} 
    \begin{algorithmic}[1]
        \REQUIRE $\Xi$: Planning task, $F_\Xi$: Diversity features, $k$: Required plans count
        \ENSURE $\Psi_\Xi$: Set of plans with different behaviours, $BC$: Behaviour count. 
        \STATE $\Psi_\Xi \gets \emptyset; BC\gets 0$\label{alg-line:bspace-planner-initial-set}
        \STATE $\textbf{do}$ \label{alg-line:fbi-main-loop-start}
        \STATE $\quad \pi \gets \operatorname{BehaviourGenerator}_\texttt{X}(\Xi, F_\Xi, \Psi_\Xi)$ 
        \STATE $\quad \Psi_\Xi \gets \Psi_\Xi \cup \{\pi\}$ 
        \STATE $\textbf{while }|\Psi_\Xi|< k \textbf{ and } \pi \not = \emptyset$ \label{alg-line:fbi-main-loop-end}
        \STATE $BC \gets \vert\Psi_\Xi\vert$ 
        \RETURN $\Psi_\Xi, BC$
    \end{algorithmic}
\end{algorithm}

\textbf{Iterated Width Planner.} 
There are only very few attempts to support planning with simulators, which  suggest using an Iterated Width (\texttt{IW(i)})~\cite{lipovetzky2015classical} planning approach. 
This is a planner that explores promising states via a sequence of calls to the simulator, each a standard breadth-first search (BFS) that prunes states when they fail to make true a new tuple of $i$ predicates. For instance, when $i=1$, the planner checks if it can construct a set of one new predicate that has been set to true for the first time, starting from the root node and progressing to the node being checked. Similarly, when $i=2$, the planner checks if it can construct a set of two new predicates, and if not the state is eliminated. This is called novelty pruning. The process continues   until $i$ reaches an upper bound value $N$ or the goal state has been reached. 
 
\begin{figure}
    \centering
    \includegraphics[scale=0.8,center]{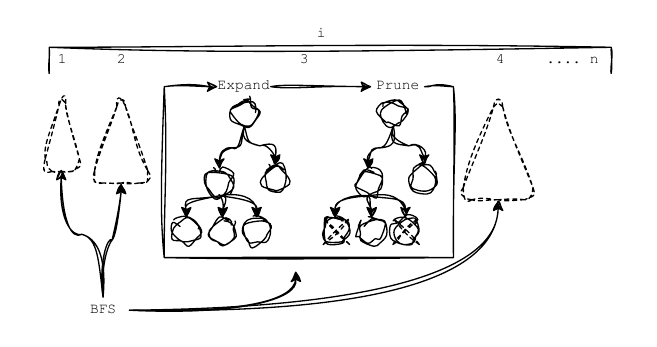}
    \caption{\texttt{IW(i)} operation. Dotted trees indicate a breadth-first search, and \texttt{i} denotes the novelty pruning threshold. Dotted crosses indicate nodes that failed the novelty check.}
    \label{fig:iw-k-operation}
\end{figure}

\Cref{fig:iw-k-operation} illustrates this operation. \texttt{IW} starts with $i=1$ and performs a breadth-first search. If it cannot find a plan, then \texttt{IW} increments $i$ until it finds a plan or $i$ meets a terminating value. During node expansion, its successor nodes are checked against the novelty criterion and are removed if they fail this check. 
In our work, this criterion is represented by a function called $\operatorname{IsNovel}:\{S^\ast\}\times\mathbb{N}\rightarrow\{\top,\bot\}$, which takes a sequence of states (e.g., $s_0,s_1,\ldots,s_n$) and a number of predicates $i$ as input and returns true if in the last state $s_n$ a new set of $i$ predicates are generated and not generated in $s_0,s_1,\ldots,s_{n-1}$, otherwise returns false. 

\textbf{Linear Temporal Logic.} \citet{pnueli1977temporal} introduced a formal specification language for verifying computer programs called Linear Temporal Logic. As a result, \texttt{LTL} has been widely used to formally describe complex properties over time in various applications and contexts, including formal methods~\cite{rozier2011linear} and Artificial Intelligence~\cite{systems11110535}. 
An \texttt{LTL} formula $\varphi$ is defined over a non-empty set of propositional symbols $\mathcal{P}$. It is closed under the Boolean connectives ($\wedge, \vee, \neg$), the binary temporal operator $\operatorname{until}(\operatorname{U})$ and the unary temporal operator $\operatorname{next-time}(\bigcirc)$. The $\operatorname{eventually}(\lozenge)$, $\operatorname{always}(\square)$, and $\operatorname{release}(\operatorname{R})$ are other temporal operators that are used to express time relations between \texttt{LTL} formulas. 
Formally, the grammar for an \texttt{LTL} formula over a given set of propositions $\mathcal{P}$ is:
$\varphi := p \vert \neg\varphi \vert \varphi_a \wedge \varphi_b \vert \varphi_a\operatorname{U}\varphi_b \vert 
\bigcirc\varphi \vert \lozenge\varphi \vert \square \varphi \vert \varphi_a \operatorname{R} \varphi_b \vert \top \vert \perp$, where $p\in\mathcal{P}$ and $\varphi_{a,b}$ are \texttt{LTL} formulas.
%
In this work, \texttt{LTL} formulas are evaluated on \emph{finite traces} and following the grammar mentioned earlier. A trace $(s_0, s_1, \ldots, s_n)$ consists of a finite sequence of states ordered linearly, where each state $s \in S$. 


\section{Behaviour Planning with Simulators}\label{sec:realisation-behaviour-planning}
In this section, we present an implementation of behaviour planning to address simulation-based planning challenges. This approach replaces the declarative model of the planning problem’s dynamics with a simulator that accepts a state and an action, and outputs a new state after applying the action to the given state. This implementation modifies the \texttt{IW(i)} to become a diverse planner, utilising Linear Temporal Logic (\texttt{LTL}) to represent the diversity of plans. 
To ensure the generation of new behaviours, we must eliminate parts of the search tree that would lead to the discovery of a solution with a behaviour previously generated. One approach to achieve this is through the use of path constraints~\cite{edelkamp2011heuristic}. Path constraints validate whether a sequence of states adheres to a specified set of constraints. If the sequence complies with the constraints, the path is not pruned from the tree; otherwise, it is removed. Path constraints are typically modelled using \texttt{LTL} logic~\cite{edelkamp2011heuristic}, which serves as the rationale behind our adoption of \texttt{LTL} to represent the behaviour space.



\subsection{Behaviour Space}
We represent a behaviour as a conjunction of \texttt{LTL} formulas, where each formula represents a dimension. We use $\mathcal{LT\!L}(A)$ 
to refer to a set of all possible combinations of \texttt{LTL} formulas for a set of boolean variables $A$. 
Two behaviours are considered the same if they have the same \texttt{LTL} formulas.
The extract function $\odot_{i}:\Pi_\Xi\rightarrow\mathcal{LT\!L}(\Delta_{i})$ returns an \texttt{LTL} formula representing the plan ($\pi$) for feature $f_i$.
In this work, we implement two features from the five proposed by \citet{abdelwahed2024behaviour} to represent diversity: cost bound ($f_{cb}$) and goal predicate ordering ($f_{go}$). 

\textbf{Feature I - Cost Bound ($f_{cb}$).} This feature differentiates between plans based on their cost. 
The corresponding cost-bound domain ($\Delta_{cb}$) comprises a set of boolean variables representing the values between 0 and the cost bound and a goal-state variable to indicate if the state is a goal state or not ($c$) (i.e., $\Delta_{cb} = \{\texttt{cost-0},\ldots,\texttt{cost-c}, \texttt{goal-state}\}$). The \texttt{LTL} formula for this dimension is $\lozenge\square(\texttt{cost-X} \wedge \texttt{goal-state})$, which means that once a cost reaches value \texttt{X} at the goal state, it should not change. 
%

\textbf{Feature II - Goal predicate ordering ($f_{go}$).} 
This feature considers the overall order in which goal Boolean variables are fulfilled and distinguishes between plans based on this order.
The corresponding domain ($\Delta_{go}$) comprises a set of \texttt{LTL} formulas 
indicating the total ordering of goals' boolean variables. The \texttt{LTL} formula for this dimension is $(\neg \texttt{g}_b\operatorname{U}\texttt{g}_a)$, which means that $\texttt{g}_a$ should be achieved before $\texttt{g}_b$. 
%
There are some planning tasks where the planner is forced to undo a goal predicate and then redo it to solve the planning task. In this case, we use helper predicates where each goal predicate has another copy set to be true the first time and to remain true. 

\textbf{Behaviour extraction.} Behaviour space is the conjunction of all dimensions' \texttt{LTL} formulas defined in the behaviour space $BS_{\Delta}$. Given a plan $\pi$ and a set of extracting functions $\odot_{\Delta}$, we then extract the plan behaviour $\mathcal{B}$ using  $\operatorname{PBehaviour}(\odot_\Delta, \pi)$ as described above. 

\subsection{Forbid Behaviour $\text{Iterative}_\texttt{LTL}$}

To realise \fbits, we need to implement the $\operatorname{BehaviourGenerator}$ used in \Cref{alg:fbi-planner-main}. This function inputs a diversity planning task $\Xi$, the diversity features $F_\Xi$, and a set of behaviours $\Psi_\Xi$. It returns a plan with a behaviour not in $\Psi_\Xi$ (i.e., $\operatorname{PBehaviour}(\odot_\Delta,\pi^\prime)\not\in \{\operatorname{PBehaviour}(\odot_\Delta,\pi)\vert\pi\in\Psi_\Xi\}$) or an empty set if it fails to find a new behaviour.
%
\texttt{IW(i)} is a promising planner with simulators since it competes with best-first search planners using advanced heuristics, making it the best candidate to use for $\operatorname{BehaviourGenerator}$. 
To transform \texttt{IW(i)} into a diverse planner, we 
 extend the its operation   to eliminate discovered behaviours. A behaviour is represented as a conjunction of \texttt{LTL} formulas, which is subsequently utilised to prune nodes in the search tree  that satisfies it.
\Cref{alg:solver-behaviour} employs a breadth-first search to explore the planning space while enforcing novelty and behaviour constraints. It starts from the initial state and applies actions to expand successor states using the problem's simulator (Lines \ref{alg:iw-diverse-initial-state}-\ref{alg:iw-diverse-explore-states}). The simulator is implemented using a function called $\operatorname{simulate}:S\times A\rightarrow S$. Following \texttt{IW(i)},  the  successor states generated can only be explored if they meet three conditions: (i) the resulting state is novel according to a novelty test, (ii) the behaviour induced by the current path is not a discovered behaviour, and (iii) the state has not been visited before (Line \ref{alg:iw-diverse-conditions-start}). The corresponding action sequence is returned as a valid plan if a goal state is reached under these constraints (Lines \ref{alg:iw-diverse-goal-check-start}-\ref{alg:iw-diverse-goal-check-end}). Otherwise, the algorithm continues exploring until no states remain to explore, returning the empty plan (Line \ref{alg:iw-diverse-emptyset-return}). The condition $b\not\in \mathit{behaviours}$ guarantees the generation of semantically distinct plans because it compares plans based on diversity features rather than actions or states.

\begin{algorithm}
\caption{$\operatorname{BehaviourGenerator}_\texttt{X}$}\label{alg:solver-behaviour}
    \begin{algorithmic}[1]
        \REQUIRE $\Xi$: Planning task, $F_\Xi$: Diversity Features, $\Psi_\Xi$: Set of plans with different behaviours
        \ENSURE A plan ($\pi$) not in $\Psi_\Xi$ or an empty set ($\emptyset$) if no plan is found.
        \STATE $N\gets \mathit{Max_N}$ \hfill{$\text{$\mathit{Max_N}$ is upper bound for \texttt{IW}}$}
        \STATE $\odot_{\Delta}\gets \{\odot \vert \langle\Delta,\odot\rangle \in F_\Xi\}$ 
        \STATE $\mathit{behaviours} \gets \{\bigwedge_{l\in\operatorname{PBehaviour}(\odot_\Delta,\pi)} l\vert\pi\in\Psi_\Xi\} $
        \FOR{$i\in [1\dots N]$}
            \STATE $Q \gets [([I], [])]$ \label{alg:iw-diverse-initial-state}
            \STATE $\mathit{Visited} \gets \{I\}$
            \WHILE{$Q$ is not empty}
                \STATE $s, path \gets \operatorname{dequeue}(Q)$
                \FOR{each $a \in A$} 
                    \STATE $s^\prime \gets s + \operatorname{simulate}(\operatorname{last}(s),a)$ \label{alg:iw-diverse-explore-states} 
                    \STATE $\pi^\prime \gets [a]+path$
                    \STATE $b\gets \bigwedge_{l\in\operatorname{PBehaviour} (\odot_{\Delta}, \pi^\prime)} l$
                    \IF{$\operatorname{IsNovel}(s^\prime, i)$ \label{alg:iw-diverse-conditions-start} \\
                    \hspace{1.5em} $\land\ b \not\in \mathit{behaviours} $ \\
                    \hspace{1.5em} $\land\ \operatorname{last}(s^\prime) \notin \mathit{Visited}$} \label{alg:iw-diverse-conditions-end}
                        \STATE $\mathit{Visited} \gets \mathit{Visited} \cup \{\operatorname{last}(s^\prime)\}$
                        \STATE $\operatorname{enqueue}(Q, (s^\prime, \pi^\prime))$
                        \IF{$\operatorname{last}(s^\prime) == G $} \label{alg:iw-diverse-goal-check-start}
                            \RETURN $\pi^\prime$
                        \ENDIF \label{alg:iw-diverse-goal-check-end}
                    \ENDIF
                \ENDFOR
            \ENDWHILE
        \ENDFOR
        \RETURN $\emptyset$ \label{alg:iw-diverse-emptyset-return}
    \end{algorithmic}
\end{algorithm}

The $\operatorname{PlanGenerator}$ should be implemented to handle the scenario where the required number of plans exceeds the number of available behaviours (\Cref{alg:plan-generator}). It follows the same steps as the $\operatorname{BehaviourGenerator}$ (\Cref{alg:solver-behaviour}) except that it rejects plans that already exist in $\Psi_\Xi$.

 \begin{figure}
    \centering
    \includegraphics[scale=0.35,center]{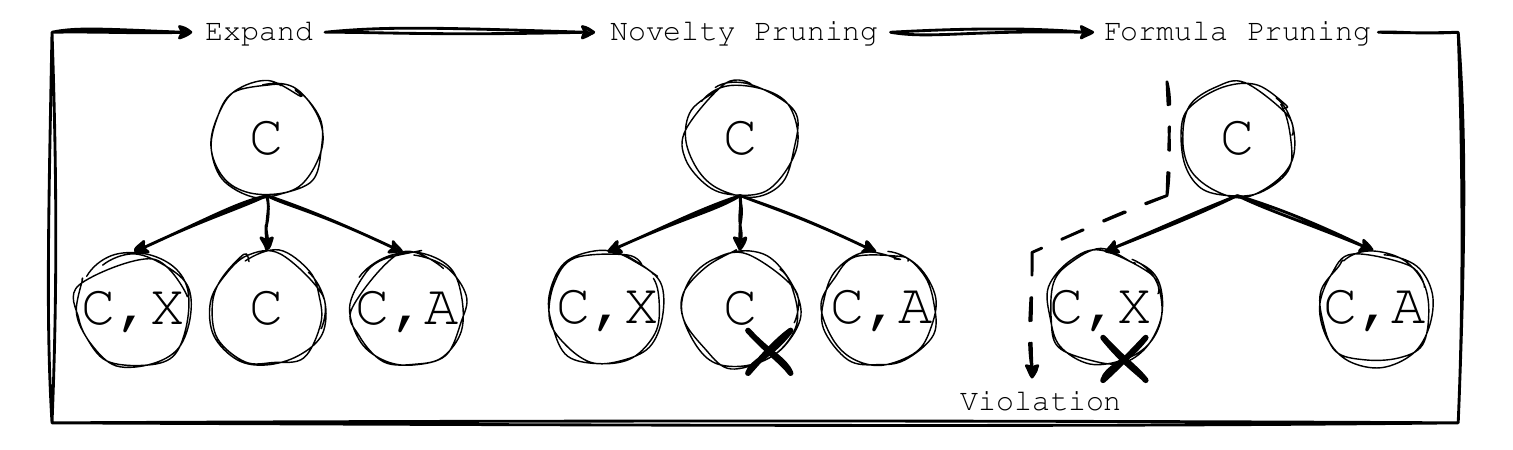}
    \caption{The extended Iterated Width BFS search filters nodes that fail the novelty test or violate a given \texttt{LTL} formula. The cross represents pruned nodes, while letters represent boolean variables set to true in the current state.}
    \label{fig:modified-iwk-operation}
\end{figure}

\Cref{fig:modified-iwk-operation} illustrates the \fbits{} operation, which rejects plans with behaviour $b_1$ that achieves boolean predicates $C$ then $X$. Two nodes are pruned. The first node is pruned due to the novelty check because no new predicates are encountered in the trace (i.e., from the root node to the middle leaf node). The second node is pruned because its behaviour is $b_1$. The node on the right in the formula pruning phase is not pruned because its behaviour is different from $b_1$. This loop continues until all behaviours are identified or all nodes are pruned during the novelty pruning phase. 

\begin{algorithm}[!t]
\caption{$\operatorname{PlanGenerator}_\texttt{X}$}\label{alg:plan-generator}
    \begin{algorithmic}[1]
        \REQUIRE $\Xi$: Planning task, $\Psi_\Xi$: Set of plans with different behaviours
        \ENSURE A plan ($\pi$) not in $\Psi_\Xi$ or an empty set ($\emptyset$) if no plan is found.
        \STATE $\text{; MAX-N is upper bound for \texttt{IW}}$
        \STATE $N\gets Max_N$
        \FOR{$i\in [1\dots N]$}
            \STATE $Q \gets [([I], [])]$
            \STATE $Visited \gets \{I\}$
            \WHILE{$Q$ is not empty}
                \STATE $s, path \gets \operatorname{dequeue}(Q)$
                \FOR{each $a \in A$} 
                    \STATE $s^\prime \gets s + \operatorname{simulate}(\operatorname{last}(s),a)$
                    \STATE $\pi^\prime \gets [a]+path$
                    \IF{$\operatorname{IsNovel}(s^\prime, i)$ \\
                    \hspace{1.5em} $\land\ \pi^\prime \not\in \Psi_\Xi$ \\
                    \hspace{1.5em} $\land\ \operatorname{last}(s^\prime) \notin Visited$}
                        \STATE $Visited \gets Visited \cup \{\operatorname{last}(s^\prime)\}$
                        \STATE $\operatorname{enqueue}(Q, (s^\prime, \pi^\prime))$
                        \IF{$\operatorname{last}(s^\prime) == G $}
                            \RETURN $\pi^\prime$
                        \ENDIF
                    \ENDIF
                \ENDFOR
            \ENDWHILE
        \ENDFOR
        \RETURN $\emptyset$
    \end{algorithmic}
\end{algorithm}



\section{Experimental Setup, Evaluation \& Findings}\label{sec:exp-diss}
%
%
This section presents experiments to evaluate our behaviour planning approach for generating diverse plans using simulators. We describe the experimental setup,  and results, followed by an analysis of our framework's limitations and potential improvements.
Our research question is: \textbf{How effectively can our behaviour planning approach generate diverse plans when working with simulators?}

\subsection{Experiment Setup}

We implemented our behaviour planning approach using Python\footnote{The code will be publicly available upon publication.}.
We integrated the FLLOAT library\footnote{https://github.com/whitemech/flloat} as the \texttt{LTL} formula satisfiability checker. Since benchmark datasets for planning with simulator problems are currently unavailable~\cite{benke2023diverse}, we evaluated our approach using PDDLGym~\cite{pddlgym} and two case studies we introduced: Puzznic and Network Penetration Testing. These domains collectively offer comprehensive test cases and contribute new benchmarks to the research community.

%

%
%
%
The experimental configuration was as follows: we solved planning tasks on an AMD EPYC 7763 64-Core Processor (2.4 GHz), generating $k$ plans where $k \in \{2,5,10,100\}$ and computing the behaviour count for each set of plans. Resource constraints were set to one CPU core, 30 minutes, and 16 GB of memory per task, with each task solved once.
Since no existing diverse planners support planning with simulators, we compared our full approach (\fbits{}) against a simplified baseline version (\fbitsnaive{}) that only forbids plans without actively promoting behavioural diversity. \fbitsnaive{} is a simple top-k planner that maintains the frontier and keeps accumulating found plans.
A t-test to assess statistical significance is computed between \fbitsnaive{} and \fbits{} using a vector of behaviour count pairs per commonly solved instance per $k$.
%


\subsection{Benchmark Problems}

\textbf{PDDLGym}~\cite{pddlgym} is a Python framework that automatically constructs OpenAI Gym~\cite{openai-gym} environments from PDDL domains and problems.
This library provides an ideal testing platform for our approach by hiding the actions' structure from the planner and offering simulator-based environments. We evaluated our method across 61 domains encompassing a total of 1290 instances, providing an extensive set of instances to validate the generalisability of our approach. This dataset has planning instances count from 1 instance per domain up to 50 instances per domain.
We configured our experiments using two general features: the goal predicate ordering ($f_{go}$) and the cost bound ($f_{cb}$), with a cost bound of 1000 actions. The state is represented using a set of boolean predicates reflecting the PDDL's problem predicates defined in the problem.

\begin{figure}
    \centering
    \includegraphics[scale=0.3, center]{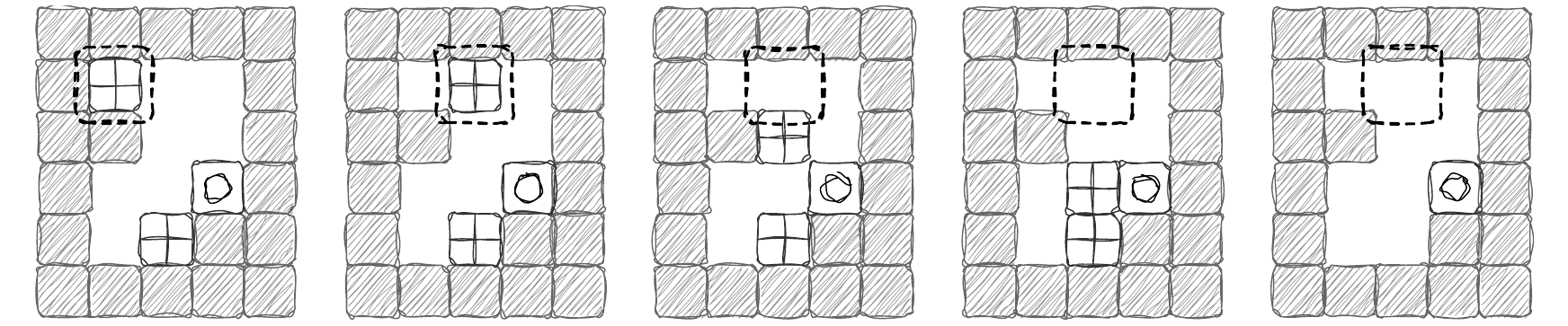}
    \caption{The Puzznic game's dynamics are as follows: the cursor moves across the box to the left, and gravity is applied since the box has no objects below it. Once the falling box touches the matched boxes, they are cleared.}\label{fig:puzznic-dynamics}
\end{figure}

\textbf{Puzznic} is a tile-matching video game published by Taito in 1989. \Cref{fig:puzznic-dynamics} illustrates the game dynamics. In Puzznic, players manipulate blocks in a grid where adjacent blocks of identical patterns match and disappear when grouped together. The blocks are affected by gravity, resulting in complex state changes and cascading matches when blocks fall and create new matching opportunities. The goal is to clear all blocks in a level, with scores based on the size and number of cascading matches achieved.
Given the inherent determinism, the game can be seen as a planning problem: given an environment (grid, blocks, and dynamics), the task is to find a sequence of actions that transforms the initial state to the goal state (all blocks cleared) while respecting the game's constraints.
%
%
Recent work~\cite{espasa2024cross} has modelled a simplified version of the game in PDDL. Gravity was modelled using derived predicates, which eased the representation of complex falling dynamics but limited compatibility to only planners that support derived predicates. The model's complexity made it challenging for planners to solve many instances. This makes Puzznic a suitable and challenging benchmark domain. By using a simulator (e.g., Python game implementation) instead, we are able to abstract game dynamics such as gravity. 
%
The level designer of Puzznic might be interested in exploring the diverse ways a level can be solved, as this could enhance the game’s replayability. Thus, the level designer would use an agent to generate a set of diverse solutions for a given level to assess its diversity. 
For Puzznic, in our experiments we solved the first 50 levels from the original game\footnote{\url{https://gamefaqs.gamespot.com/gameboy/579654-puzznic/faqs/65879}}. We distinguish plans based on two features: the order in which blocks are cleared and the final score achieved when solving a level. The first feature is captured by the goal predicate ordering ($f_{go}$) while the second one is captured by the cost bound feature ($f_{cb}$). The score is computed at the end of the level and is represented with a boolean variable. We bound the cost to 1000 moves. As for the state representation, we represent the cursor, boxes positions, another variable to indicate if the state is a goal state or not and another variable to reflect the score.


\begin{figure*}
    \centering
    \includegraphics[scale=0.35,center]{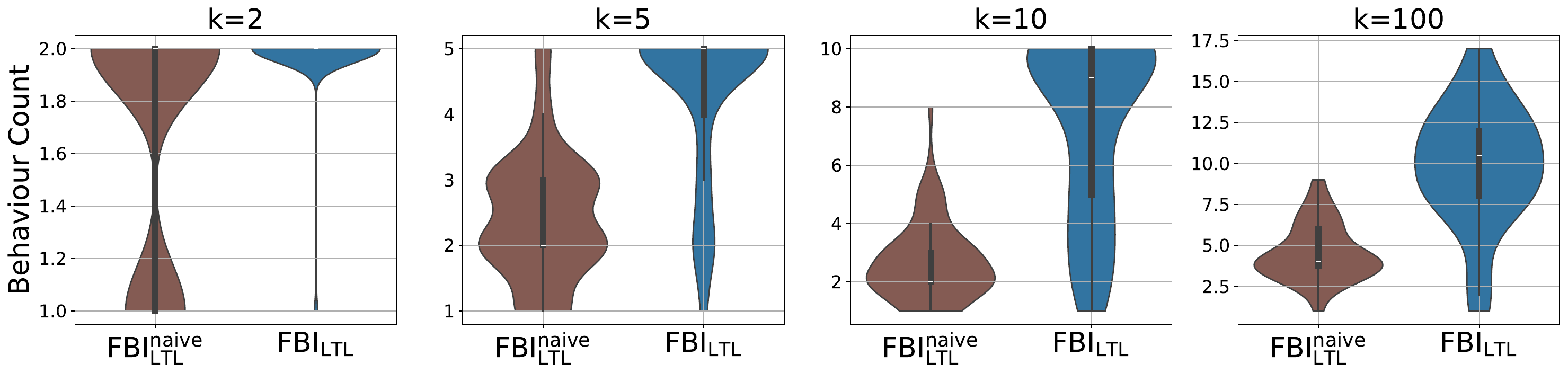}
    \caption{Violin plots for all planning problems experiments. Planners are ordered as follows: \fbits, \fbitsnaive.}\label{fig:violin-results-ltl}
\end{figure*}

\textbf{Network Penetration Testing} helps uncover potential attack vectors on an organisation's network. By identifying all possible methods an attacker might employ to infiltrate the network, security experts can proactively reinforce security measures, significantly enhancing the network's resilience against real-world cyber threats. In this context, an agent  equipped with a diverse planner can help discover vulnerabilities in a network. 
Cybersecurity experts can leverage the diverse plans to improve the network defences and enhance their resilience by identifying multiple valid attack plans that can exploit their network's vulnerabilities.
%
We employ NASim, a simulator to model detailed network configurations (e.g., network defences) and simulate attacks in a controlled, declarative manner~\cite{pengym}.
Here, we used all 18 instances from NASim. We distinguish plans based on the goal predicate ordering feature ($f_{go}$), which determines the order in which the compromised hosts are 
addressed. 
Regarding the state representation, NASim represents a state as a matrix of connected hosts. We transformed this matrix into a set of boolean predicates that represent the value of each entry in the matrix, along with another predicate that indicates which hosts are compromised at a specific state.

\begingroup
\setlength{\tabcolsep}{1.2pt}
\renewcommand{\arraystretch}{1.1} 
\begin{table}[b]
\caption{Experimental results comparing \fbits{} and \fbitsnaive{} across different values of $k$. Coverage shows the number of solved instances for each approach. CI indicates the number of commonly solved instances used for comparison. BC presents the accumulated behaviour counts, with bold values indicating statistically significant differences ($p < 0.05$) based on t-tests. Execution time shows the average solving time in minutes per planning problem.}\label{tbl:all-results-ltl}
\small
\centering
\begin{tabular}{c|cc|c|cc|cc}
\hline
\multirow{2}{*}{k} & \multicolumn{2}{c|}{Coverage} & \multirow{2}{*}{CI} & \multicolumn{2}{c|}{BC}      & \multicolumn{2}{c}{Exec. Time [mins]} \\
                   & \fbits      & \fbitsnaive     &                     & \fbits         & \fbitsnaive & \fbits          & \fbitsnaive         \\ \hline
2                  & 193         & 222             & 142                 & $\textbf{281}$ & 239         & 2.85            & 1.21                \\
5                  & 108         & 201             & 104                 & $\textbf{452}$ & 252         & 4.83            & 0.47                \\
10                 & 78          & 179             & 73                  & $\textbf{537}$ & 195         & 8.17            & 0.40                \\
100                & 33          & 97              & 32                  & $\textbf{317}$ & 146         & 12.41           & 0.51                \\ \hline
\end{tabular}

\end{table}

\endgroup

\begingroup
\setlength{\tabcolsep}{1.0pt}
\renewcommand{\arraystretch}{1.1} 
\begin{table}[t]
\caption{Behaviour count values for some selected domains. The table follows the same format as \Cref{tbl:all-results-ltl}. All zero rows are omitted. }\label{tbl:some-results-ltl}
\small
\centering
\begin{tabular}{c|c|cc|c|cc}
\hline
\multirow{2}{*}{Domain}                                      & \multirow{2}{*}{k} & \multicolumn{2}{c|}{Coverage} & \multirow{2}{*}{CI} & \multicolumn{2}{c}{BC} \\
                                                    &     & \fbits & \fbitsnaive &    & \fbits         & \fbitsnaive \\ \hline
\multirow{3}{*}{Puzznic}                            & 2   & 9      & 9           & 9  & 17             & 14          \\
                                                    & 5   & 5      & 6           & 5  & $\textbf{23}$  & 10          \\
                                                    & 10  & 3      & 5           & 3  & $\textbf{24}$  & 4           \\ \hline
\multirow{4}{*}{\makecell{Network\\Security}}         & 2   & 13     & 12          & 10 & $\textbf{18}$  & 10          \\
                                                    & 5   & 8      & 12          & 8  & $\textbf{13}$  & 8           \\
                                                    & 10  & 6      & 10          & 6  & $\textbf{9}$   & 6           \\
                                                    & 100 & 3      & 5           & 2  & 4              & 3           \\ \hline
\multirow{4}{*}{\makecell{PDDLEnv\\Trap\\newspapers}} & 2   & 4      & 4           & 4  & 8              & 6           \\
                                                    & 5   & 4      & 4           & 4  & $\textbf{20}$  & 6           \\
                                                    & 10  & 3      & 4           & 3  & $\textbf{26}$  & 6           \\
                                                    & 100 & 3      & 4           & 3  & $\textbf{30}$  & 10          \\ \hline
\multirow{4}{*}{\makecell{PDDLEnv\\Minecraft}}        & 2   & 28     & 28          & 28 & $\textbf{56}$  & 45          \\
                                                    & 5   & 28     & 28          & 28 & $\textbf{140}$ & 66          \\
                                                    & 10  & 25     & 28          & 24 & $\textbf{237}$ & 70          \\
                                                    & 100 & 15     & 24          & 15 & $\textbf{163}$ & 70          \\ \hline
\multirow{4}{*}{\makecell{PDDLEnv\\Easy\\newspapers}} & 2   & 19     & 22          & 19 & 38             & 38          \\
                                                    & 5   & 14     & 22          & 14 & $\textbf{70}$  & 39          \\
                                                    & 10  & 11     & 22          & 11 & $\textbf{100}$ & 26          \\
                                                    & 100 & 7      & 21          & 7  & $\textbf{76}$  & 36          \\ \hline
\multirow{3}{*}{\makecell{PDDLEnv\\Manylogistics}}    & 2   & 6      & 7           & 6  & $\textbf{12}$  & 7           \\
                                                    & 5   & 5      & 6           & 5  & $\textbf{25}$  & 12          \\
                                                    & 10  & 2      & 6           & 2  & $\textbf{20}$  & 4           \\ \hline
\multirow{4}{*}{\makecell{PDDLEnv\\Search\\And\\RescueLevel4}} & 2                  & 28            & 41            & 28                  & $\textbf{56}$   & 51   \\
                                                    & 5   & 24     & 40          & 24 & $\textbf{90}$  & 65          \\
                                                    & 10  & 17     & 39          & 17 & $\textbf{77}$  & 59          \\
                                                    & 100 & 2      & 26          & 2  & 9              & 7           \\ \hline
\end{tabular}
\end{table}
\endgroup

\subsection{Results and Discussion}


\begin{figure*}
    \centering
    \includegraphics[width=0.8\linewidth]{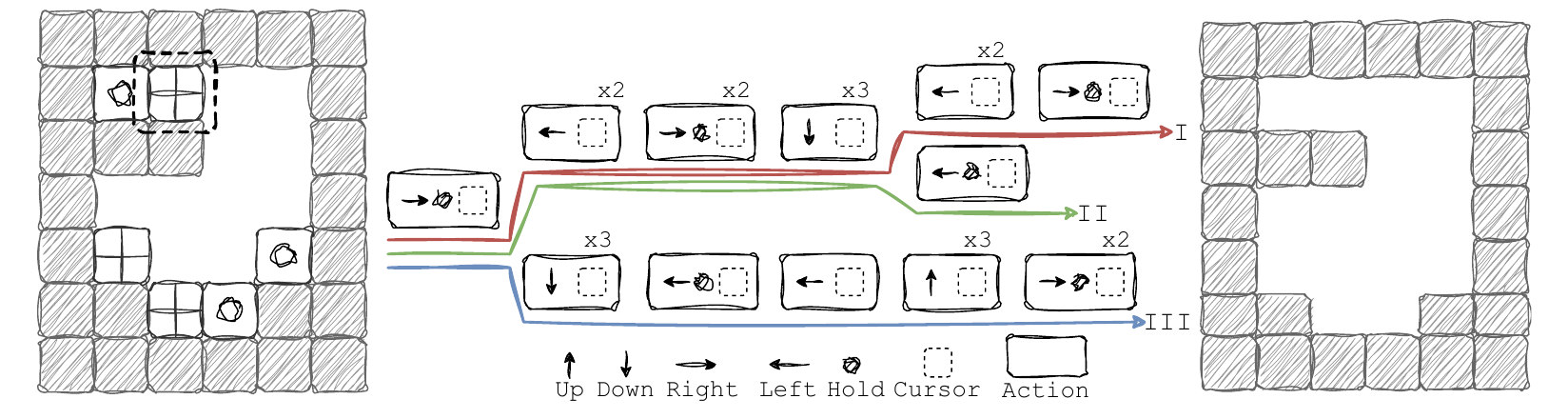}
    \caption{Three plans for solving Puzznic's first level. Check supplementary material for the plans' actions.}
    \label{fig:puzznic-example}
\end{figure*}

\begin{figure}
    \centering
    \includegraphics[scale=0.9]{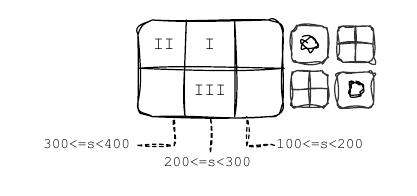}
    \caption{Behaviour space for Puzznic: A 2D grid to represent diversity model. Plans are distinguished based on two factors. The first feature is the order in which the boxes were cleared. The second feature is the final score (s) of the plan. Letters \texttt{I}, \texttt{II} and \texttt{III} represent the plans displayed in \Cref{fig:puzznic-example}.}
    \label{fig:puzznic-example-bspace}
\end{figure}

\Cref{tbl:all-results-ltl} presents the accumulated coverage, behaviour count and execution time for the commonly solved instances by the planners. We use Violin plots to visualise the distribution, central tendency and spread of the behaviour count values for the accumulated domains results, this is presented in \Cref{fig:violin-results-ltl}. \Cref{tbl:some-results-ltl} shows the accumulated coverage and behaviour count for some selected domains by the planner. The selected domains are chosen on the coverage value, we only show the domains with coverage of at least 4 for $k=2$. Domains with coverage less than that are omitted for space reasons.  More information regarding all PDDLGym domains are provided in the supplementary materials. 
\Cref{tbl:all-results-ltl}, shows we can generate diverse plans for simulation-based planning problems, especially given the problems shown in \Cref{tbl:some-results-ltl}. The not shown one are either not solved or the \fbits{} was able to generate more diverse plans compared to its naive version.
This comes at the cost of increased execution time which keeps increasing based on $k$. This occurs because \fbits{} is searching for plans that are semantically different unlike its naive version. There is a primary factor that affects the coverage, which is the number of required plans $k$. When $k$ increases this increases the execution time since the planner spends more time in path constraints checks in order to find new behaviours which increases the chances of hitting the time-limit (30 minutes). 

The general lower coverage  of \fbits{} in comparison with \fbitsnaive{} can be attributed to two main reasons. Firstly, the base planner we are using to generate $k$ plans, Iterated Width~\cite{lipovetzky2015classical}, is a standard breadth-first search where states are pruned based on some conditions. The approach is the best suited for our needs but the absence of heuristics hurts its coverage. Secondly, producing diverse plans is also a challenge: Planning is PSPACE-Complete~\cite{Computational-classical-planning}, and finding a diverse set of solutions is NP-hard~\cite{kuo1993analyzing}. Combining the two with the blind search approach results in low coverage, as seen in Table \ref{tbl:all-results-ltl}.
To confirm this claim, we ran \fbitsnaive{} asking for only one plan ($k=1$). It successfully solved 434 out of 1358 instances (1290 PDDLGym, 50 Puzznic and 18 Network Penetration Testing instances). This indicates that the planning problems we used are already challenging to solve. When we applied the same planner to generate two plans ($k=2$), the number of successful solutions dropped to 225. Note that the increase in $k$ does not necessarily translate into a corresponding increase in the behaviour count, as the behaviour count is influenced by the specific planning problem. For instance, the PDDL domains presented in \Cref{tbl:some-results-ltl} demonstrate that when $k=100$, \fbits{} will not generate 100 behaviours for a planning task if the task itself does not have 100 behaviours available.
%
%
We chose the PDDLGym library as a benchmarking library because there was not a specialised library designed for this type of planning problem. Tables~\ref{tbl:all-results-ltl} and \ref{tbl:some-results-ltl} show that \fbits{} is able to generate diverse plans for such setup compared to its naive version \fbitsnaive{}.
%
%
The Network security and Puzznic domains exemplify scenarios where diverse planning with simulators are highly useful. Regarding cybersecurity, current conflicts are increasingly fought in cyberspace rather than solely on land~\cite{cyberwarefare}.  Therefore, employing diverse planners for cybersecurity applications would be beneficial in developing several cost-effective defensive actions against the adversary activities~\cite{cyber-resilience}. 
%
%
Puzznic illustrates the application of diverse planning in assessing the replayability of video game levels or as a starting point for automated level design. According to \citet{game-diversity-measure}, one crucial aspect that significantly influences player engagement and satisfaction is the diversity of game content. \citeauthor{game-diversity-measure} proposed capturing the player's interaction with the game (i.e., how the players navigate and engage with the game content) to quantify a game's diversity. They quantify diversity using a distance function which is computed based on the player's sequence of actions. A more intuitive approach is to construct a behaviour space using the diversity features of interests and then use \fbits{} to evaluate the diversity of the produced content. To illustrate this, \Cref{fig:puzznic-example} presents three distinct plans that successfully solve Puzznic's initial level, along with the behaviour space employed to generate these plans (\Cref{fig:puzznic-example-bspace}). In this scenario, we differentiated between plans based on two criteria: (i) the final score achieved and (ii) the sequence of cleared boxes. \fbits{} discovered three distinct solutions to the level, including a plan that scored between 300 and 400 points and two plans that scored between 200 and 300 points, albeit with different orders of cleared boxes.

\begin{figure*}
    \centering
    \includegraphics[width=0.9\linewidth]{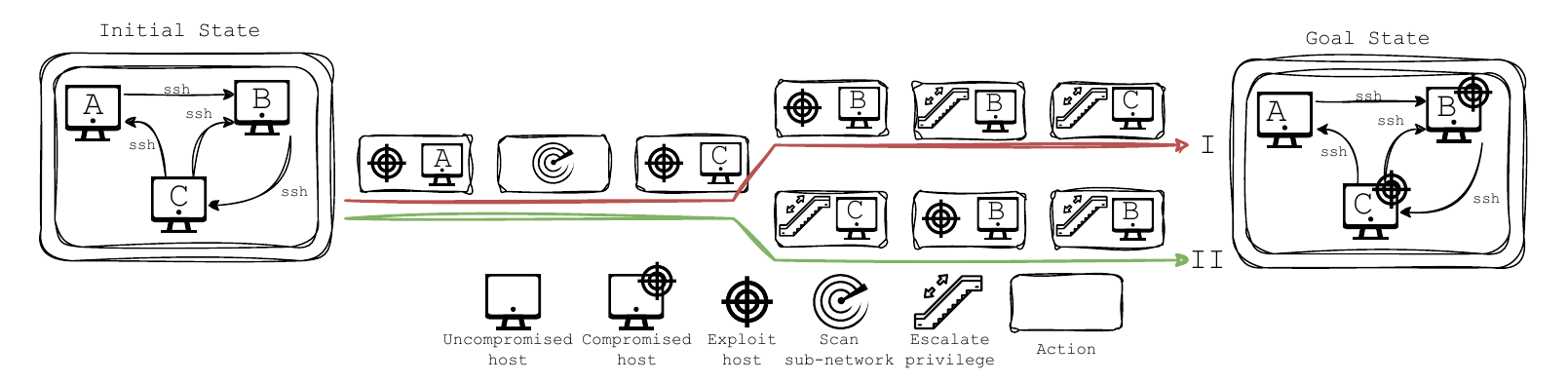}
    \caption{Two plans for solving network security, first instance, that represent an organization with three connected computers. Check supplementary material for the plans' actions.}
    \label{fig:network-security-app}
\end{figure*}


\Cref{fig:network-security-app} illustrated that diverse planning can be highly beneficial for network security applications. \fbits{} successfully identified two distinct plans to exploit an organization’s network consisting of three connected computers with a goal to exploit all three computers. Given the numerous avenues for network exploitation, it is crucial for operational security to be aware of these methods. By deploying defences against these avenues, organisations can enhance their network’s resilience to cyber attacks.
 
%

In contrast, model-based planners have higher coverage because they utilise a preprocessing phase that reduces actions based on reachability analysis and employs heuristic functions to guide the search
\cite{fastdownward}.
Beyond coverage, we also faced a challenge in modelling numeric features using \texttt{LTL}. To model such features, we use Boolean variables to represent a value. For instance, in the cost-bound feature ($f_{cb}$), each Boolean variable represents a plan cost, and it is set to true when the plan's cost reaches this value. This approach makes it challenging to perform comparison operations such as less/greater than, which can be useful when searching for plans that satisfy a specific value range.

\section{Related work} \label{sec:related-work}

\citet{benke2023diverse} suggested the first approach for generating diverse plans for planning using simulators. Their approach involved constructing a tree using any tree search algorithm and extracting diverse plans from this pre-generated tree. They formulated the planning problem as a Markov Decision Process (\texttt{MDP}), deviating from the traditional classical planning approach. This choice was made because they focused on planning problems that provide rewards rather than explicit goals, unlike our case, where we consider planning problems with explicit goals.
However, \citeauthor{benke2023diverse} highlighted that their approach extracts plans would essentially be the same when considered semantically. 

There is other work done on diverse planning, but for model-based planners. \citet{srivastava2007domain} pioneered diverse plan generation by proposing a local search-based planner (e.g., LPG~\cite{gerevini2003planning}) with distance functions. While successful, it converged to suboptimal solutions. \citet{roberts2014evaluating} addressed this by introducing a multi-queue $A^\ast$ algorithm with optimality and diversity queues. This balanced trade-off prioritises parsimony to prevent single heuristic dominance. \citet{vadlamudi2016combinatorial} proposed a two-phase optimisation approach. The initial phase generated many plans in their bi-level optimisation framework, while the subsequent phase extracted a subset of diverse plans. This framework guaranteed cost-bounded diverse plan sets. The first phase did not require a specific planner. \cite{katz-lee-socs2023} suggested using any top-k, top-q, and diverse planners for this phase, respectively. \cite{katz2020reshaping} proposed a greedy method to extract a diverse set of plans while maximising the sum of pairwise distances between plans. 
Diverse planning has found its way into the multi-agent planning setting as well. \citet{ma-diverse-planning} utilised distance metrics from diverse planning to assess the quality of the generated solutions. They generated n different solutions for the multi-agent planning problem, evaluated their qualities, and chose the one with the highest quality.

One attempt to address the limitations of the dataset was made by \citet{lipovetzky2015classical}, who used the Arcade Learning Environment (ALE)~\cite{bellemare13arcade}. They employed the online planning setting, which permitted them to perform a look-ahead before taking an action. Their state model was based on a feature called RAM State. A state is represented as a vector of 128 variables, each capable of taking 256 values. This differs from classical planning problems, where there are no explicit goal states, but the objective in this setup is to achieve higher scores. However, there are no available datasets for classical planning problems~\cite{benke2023diverse}.
While diverse planning has been explored in model-based settings, real-world problems like autonomous game level design~\cite{level-generator}, and crisis management~\cite{flood_rl} often lack declarative models and can only be interfaced through simulators. These domains benefit from diverse planning to expose vulnerabilities, generate varied user experiences, and account for uncertain dynamics. However, there is a notable absence of literature addressing diversity in simulation-based planning. \citet{benke2023diverse} is the only known exception, relying on tree extraction rather than semantically grounded diversity. 




\section{Conclusions \& Future work}\label{sec:conclusion-future}

This paper suggests using a diverse simulation-based planner for autonomous agents to overcome the limitations inherited from using a planner that produces a single plan. This was achieved by implementing a behaviour planning framework for the simulation-based planning setting.
Unlike traditional symbolic planners, our approach directly interacts with simulators, handling complex environments that are difficult to model declaratively. Our method explicitly incorporates customisable diversity dimensions during planning which align with agents objectives and preferences. This is done by leveraging a diversity model encoded as a conjunction of \texttt{LTL} formulas and 
a tree-search planner, 
Iterated Width, to generate diverse plans. 
We demonstrated the effectiveness of our implementation across several domains.
\fbits{} generated more diverse plans than a naive baseline across different planning instances and diversity models. These findings validate the potential of behaviour planning in planning problems represented using simulators.
However, our implementation also reveals two challenges which are the scalability of solving planning with simulators and modelling numeric features. 
%
%
Therefore, our future work will explore improving planner scalability via reducing the state space and incorporating richer temporal logics to support numeric features. For example, we can explore reachability analysis similar to the ones employed by Fast Downward for the model-based planning problems~\cite{fastdownward}. Such reducing will help in reducing the search time which will improve scalability. 
%
Regarding numeric features modelling, we can use other variants of temporal logic, such as Signal Temporal Logic~\cite{stl}, which naturally handles numeric constraints and is commonly used in safety and security validation of cyber-physical systems~\cite{stl-example}. Another possible future work is constructing a standardised benchmark suite for simulation-based planning.

%

\bibliographystyle{plainnat}  
\bibliography{references}

\end{document}